\pdfoutput=1

\documentclass[11pt]{article}

\usepackage{emnlp2021}

\usepackage{times}
\usepackage{latexsym}

\usepackage[T1]{fontenc}

\usepackage[utf8]{inputenc}

\usepackage{microtype}

\usepackage{graphicx}
\usepackage[font={small}]{caption}
\usepackage{url}
\usepackage{wrapfig}
\usepackage{amssymb}
\usepackage{color,xcolor,colortbl}
\usepackage{enumitem}
\usepackage{soul}
\usepackage{amsmath}
\usepackage{algorithm}
\usepackage{algorithmic}
\usepackage{booktabs}
\usepackage{bm,bbm}
\usepackage{mathtools}
\usepackage{array}
\usepackage{multirow}
\usepackage{subcaption}
\usepackage{pbox}
\usepackage{pifont}
\usepackage{arydshln}
\usepackage{dblfloatfix}
\usepackage{relsize}

\DeclareMathOperator*{\argmax}{arg\,max\,}
\newcommand\numberthis{\addtocounter{equation}{1}\tag{\theequation}}

\definecolor{darkblue}{rgb}{0.0, 0.0, 0.55}
\setul{0.5ex}{0.3ex} 
\setulcolor{blue} 
\usepackage[T1]{fontenc}
\usepackage[scaled=.8]{beramono}

\title{Modeling Endorsement for Multi-Document Abstractive Summarization}

\author{Logan Lebanoff$^\dagger$ \,\, Bingqing Wang$^\S$ \,\, Zhe Feng$^\S$ \,\, Fei Liu$^\dagger$\\[0.8em]
$^\dagger$University of Central Florida, Orlando, FL\\
$^\S$Robert Bosch LLC, Sunnyvale, CA\\[0.8em]
\texttt{loganlebanoff@knights.ucf.edu} \quad 
\texttt{\{bingqing.wang,zhe.feng2\}@us.bosch.com}\\
\texttt{feiliu@cs.ucf.edu}
}

\date{}

\begin{document}
\maketitle
\begin{abstract}

A crucial difference between single- and multi-document summarization is how salient content manifests itself in the document(s).
While such content may appear at the beginning of a single document, essential information is frequently reiterated in a set of documents related to a particular topic, resulting in an endorsement effect that increases information salience.
In this paper, we model the cross-document endorsement effect and its utilization in multiple document summarization. 
Our method generates a synopsis from each document, which serves as an endorser to identify salient content from other documents.
Strongly endorsed text segments are used to enrich a neural encoder-decoder model to consolidate them into an abstractive summary.
The method has a great potential to learn from fewer examples to identify salient content, 
which alleviates the need for costly retraining when the set of documents is dynamically adjusted. 
Through extensive experiments on benchmark multi-document summarization datasets, we demonstrate the effectiveness of our proposed method over strong published baselines.
Finally, we shed light on future research directions and discuss broader challenges of this task using a case study.

\end{abstract}

\section{Introduction}

``Repeat a lie often enough and it becomes the truth.''
This proverb stresses the importance of \emph{repetition} and \emph{frequency} in human comprehension.
It causes an endorsement effect that increases the salience of repeated information.
In this paper, we leverage the endorsement effect to summarize multiple documents that discuss a particular event or topic (MDS). 
In the commercial arena, MDS could be used to aggregate search results~\cite{Miller:2020} and distill insights from customer reviews~\cite{brazinskas-etal-2020-shot}.
Further, MDS is an integral part of the daily work of intelligence analysts who identify important information from raw documents and consolidate it into a summary report to be disseminated to the leadership~\cite{Hamilton:2014}.

\begin{figure}[t]
\centering
\includegraphics[width=3.1in]{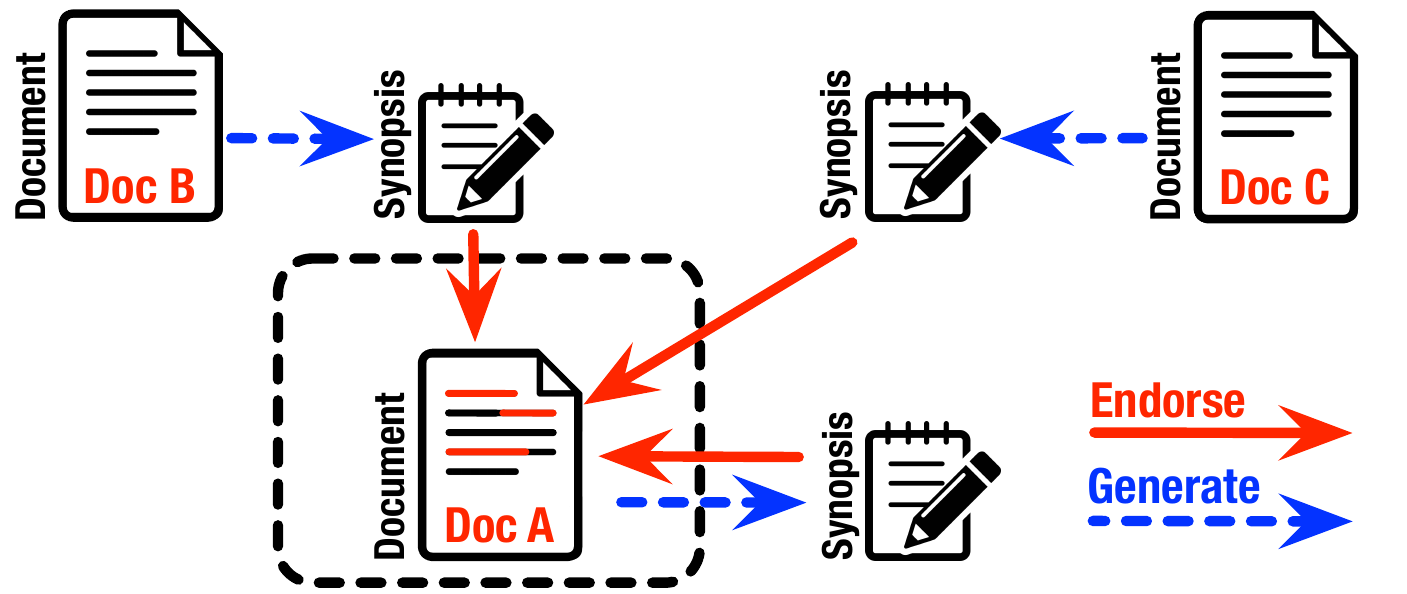}
\caption{
An example of synopsis-document relationships.
Synopsis-document endorsements are leveraged to identify important text segments from a source document (e.g., Doc A).
Strongly endorsed segments of all documents are consolidated into an abstractive summary.
}
\label{fig:framework}
\vspace{-0.1in}
\end{figure}

\textbf{Mu}lti-document \textbf{A}bstractive \textbf{S}ummarization, i.e. \textbf{MuDAS}, remains a challenging problem  compared to its single-document counterpart~\cite{see-etal-2017-get,chen-bansal-2018-fast,narayan-etal-2018-dont,JMLR:v21:20-074,lewis-etal-2020-bart}.
The task poses a substantial challenge to modern neural models:
when the set of source documents is concatenated into a flat sequence, it may exceed the maximum length allowed by the GPU memory.
There are also fewer datasets available to train MuDAS models in an end-to-end fashion.
Recent work tackles this problem by selecting representative sentences from the source documents to reduce the task to single-document summarization~\cite{lebanoff-etal-2018-adapting,coavoux-etal-2019-unsupervised,fabbri-etal-2019-multi}.

Nevertheless, there could be substantial information loss if only representative sentences are used for MuDAS.
It becomes unclear what information is reiterated and salient, resulting in unimportant sentence parts being included in the summary.
E.g., when the sentence ``\emph{World leaders join to pledge \$8 billion for vaccine, but the U.S. sits out}'' is selected from the document set, it is unclear which of its segments, ``\emph{\$8 billion}'' or ``\emph{U.S. sits out},'' is more salient given the topic of discussion.
The neural representations also treat different quantities, e.g., ``\emph{\$8 billion}'' and ``\emph{\$5 million},'' indiscriminately~\cite{rogers2020primer}.
Consequently, there is an urgent need for summarization systems to acquire fine-grained, segment-level textual salience. 
Without that, a neural abstractive system can miss out on salient details and favor fluency over information accuracy.

In this paper, we present a conceptual framework that leverages the endorsement effect to model fine-grained segment salience for multi-document summarization.
When an analyst reads a document, he retains a synopsis of the key ideas of the document in his mind.
The synopsis later serves as an endorser to identify segments in other documents that reiterate the same ideas~\cite{Hintzman:1976}.
We call the synopsis an ``\textbf{Endorser}'' and the document a ``\textbf{Candidate}.''
Segments of the candidate documents that are frequently endorsed by synopses suggest high salience and are to be consolidated into an abstractive summary. 
Our synopses are generated from a state-of-the-art summarizer~\cite{lewis-etal-2020-bart} and
a variety of methods are investigated to quantify the level of endorsement from a text synopsis to a document.
Figure~\ref{fig:framework} provides an overview of synopsis-document endorsement.

Our contributions in this paper include:
\begin{itemize}[topsep=5pt,itemsep=0pt,leftmargin=*]
\item presenting a new conceptual framework to model asynchronous endorsement from text synopses to documents for multi-document summarization;
\item devising a novel method to enrich neural encoder-decoder models with fine-grained segment-level endorsement to consolidate strongly endorsed content into an abstractive summary; and
\item through extensive experiments on multiple benchmark summarization datasets, we demonstrate the effectiveness of the endorsement method over state-of-the-art baselines. We make our code and models publicly available.\footnote{\url{https://github.com/ucfnlp/endorser-summ}}
\end{itemize}

\section{Related Work}
\label{sec:related}

Redundancy is essential in multi-document summarization. 
Without repetition and redundancy, even humans cannot agree on what information is salient and should be included in the summary~\cite{daume-iii-marcu-2004-generic}.
Optimizing summaries for frequency-based saliency has attained success prior to the era of deep learning~\cite{berg-kirkpatrick-etal-2011-jointly,Kulesza:2012,boudin-etal-2015-concept}.
These extractive systems strive to include the most frequently occurring concepts in the summary.
However, when it comes to abstractive summarization systems, the frequency of concepts is not fully utilized by modern neural models.

Recent studies on MuDAS \emph{implicitly} estimate frequency using hierarchical encoders / decoders.
Liu and Lapata~\shortcite{liu-lapata-2019-hierarchical} encode the documents using hierarchical Transformers where cross-document relationships are characterized by attention weights.
Perez-Beltrachini et al.~\shortcite{perez-beltrachini-etal-2019-generating} explore structured convolutional decoders.
Li et al.~\shortcite{li-etal-2020-leveraging} leverage similarity and discourse graphs to alter the attention mechanism of encoder-decoder models.
Researchers have also attempted optimization algorithms such as maximal margin relevance and determinantal point processes combined with contextualized representations and reinforcement learning~\cite{cho-etal-2019-improving,cho-etal-2019-multi,mao-etal-2020-multi}.
Despite promising progress, modeling frequency for multi-document summarization remains an open problem, in part because neural summarization models are often pretrained on single documents that contain little or no redundant content~\cite{kryscinski-etal-2019-neural,zhang-etal-2019-hibert,jin-wan-2020-abstractive,laban-etal-2020-summary,pmlr-v119-zhang20ae}.
Named entities and quantities that represent salient information details are not properly accounted for~\cite{xu-durrett-2021-dissecting}.
If we do not \emph{explicitly} model frequency, abstractive summarizers may fail to adequately recognize such salient details.

We are particularly interested in reducing multiple input documents to a single document, then consolidate the content into a succinct abstract~\cite{nayeem-etal-2018-abstractive,coavoux-etal-2019-unsupervised}.
Our method enhances the single document with fine-grained segment salience to offset the lead bias~\cite{grenander-etal-2019-countering,xing-etal-2021-demoting}, which hinders the development of multiple-document summarization. 
Our salience estimates are obtained from a frequency-driven endorsement model.
Below we present details of the proposed method.

\section{Summarization with Endorsement}
\label{sec:summarization}

We approach the MuDAS problem in two stages.
First, we obtain fine-grained segment-level endorsement for any candidate document.
By excluding unendorsed sentences from consideration, we reduce the set of documents to a single input document.
We next present an enhanced abstractive summarization model to consolidate the document into a succinct abstract, analogously to how an editor would consolidate text with emphasis on endorsed segments.
This process involves non-trivial design decisions. 
In this section, we start by presenting the second stage in our approach -- the summarization model with endorsement. 

\subsection{The Original Transformer}

We choose the encoder-decoder architecture over decoder-only architectures~\cite{radford2019language,NIPS2019_9464,brown2020language}.
It allows us to balance the contribution from the source text and its endorsed segments in summary generation. 
The encoder and decoder each comprise of a stack of $L$ Transformer blocks~\cite{NIPS2017_7181}.
Let $\{x\}_{i=0}^m$ be the source sequence corresponding to the input document, and $\{y\}_{j=0}^n$ the summary sequence. 
$x_0$ and $y_0$ are beginning-of-sequence symbols. 
Let $\bm{E}$ be a matrix of token embeddings and $\bm{P}$ be position embeddings.
An encoder produces a set of hidden vectors in its $l$-th layer (Eq.~(\ref{eq:encoder})), $\bm{H}^{(l)} = \langle\bm{h}_0^{(l)}, \ldots, \bm{h}_m^{(l)}\rangle$, where $\bm{h}_i^{(l)}$ is a hidden vector of the $i$-th source token.
A decoder utilizes top-layer encoder hidden vectors $\bm{H}^{(L)}$ to decode the summary sequence, where $\bm{G}^{(l)}$ represents a sequence of hidden vectors of the $l$-th decoder layer (Eq.~(\ref{eq:decoder})).
An upper triangular-shaped mask is used by the decoder, so that $\bm{g}_j^{(l)}$ only depends on summary tokens whose positions are less than $j$.
\begin{align*}
& \bm{H}^{(l)} = \langle\bm{h}_0^{(l)}, \ldots, \bm{h}_m^{(l)}\rangle\numberthis\label{eq:encoder}\\
&= \left\{
\begin{array}{ll}
\langle\bm{E}_{x_0}+\bm{P}_0, \ldots, \bm{E}_{x_m}+\bm{P}_m\rangle & l=0\\
\textsc{EncBlock}_l\big(\bm{H}^{(l-1)}\big) & l>0
\end{array}
\right.\\[0.5em]
& \bm{G}^{(l)} = \langle\bm{g}_0^{(l)}, \ldots, \bm{g}_n^{(l)}\rangle\numberthis\label{eq:decoder}\\
&= \left\{
\begin{array}{ll}
\langle\bm{E}_{y_0}+\bm{P}_0, \ldots, \bm{E}_{y_n}+\bm{P}_n\rangle & l=0\\
\textsc{DecBlock}_l\big(\bm{G}^{(l-1)}, \textcolor{black}{\bm{H}^{(L)}}\big) & l>0
\end{array}
\right.
\end{align*}

With this architecture, we argue that it is preferable to modify the decoder and cross-attention to steer it towards endorsed content, rather than modifying the encoder representations $\bm{H}^{(L)}$, as they are often unsupervisedly pretrained. 
It would be best if such representations remain unaffected by whether a segment of the source text is endorsed or not to provide model flexibility. 
A decoder layer consists of three main blocks to transform from $\bm{G}^{(l-1)}$ to $\bm{G}^{(l)}$ (Eqs.~(\ref{eq:self-attn}-\ref{eq:ffn})).\footnote{We omit the residual connection and layer normalization associated with each block for brevity.}
In particular, self-attention allows a summary token to attend to other summary tokens.
Cross-attention allows a summary token to attend to all source tokens using $\bm{H}^{(L)}$.
Finally, a feed-forward network with ReLU activation is applied to generate $\mathbf{G}^{(l)}$.
Our focus of this work is to improve the cross-attention to emphasize on endorsed content during decoding.
\begin{align*}
&\widetilde{\bm{G}}^{(l-1)} = \textsc{Self-Attn}(\bm{G}^{(l-1)})\numberthis\label{eq:self-attn}\\
&\widehat{\bm{G}}^{(l)} = \textsc{Cross-Attn}(\widetilde{\bm{G}}^{(l-1)}, \textcolor{black}{\bm{H}^{(L)}})\numberthis\label{eq:cross-attn}\\
&\mathbf{G}^{(l)} = \textsc{FeedForward}(\widehat{\bm{G}}^{(l)})\numberthis\label{eq:ffn}
\end{align*}

The \emph{original} cross-attention head $z$ transforms the $j$-th decoder state $\widetilde{\bm{g}}_j^{(l-1)}$ and $i$-th encoder state $\bm{h}_i^{(L)}$ into query, key and value vectors (Eqs.~(\ref{eq:query}-\ref{eq:value})).
It computes attention weights as a normalized dot product between query and key vectors.
The output of the head is a weighted sum of value vectors.

\subsection{Companion Heads}

We introduce a set of \emph{companion heads} for each original head. All companion heads of $z$ share the parameters $\{\bm{W}_z^Q$, $\bm{W}_z^K$, $\bm{W}_z^V\}$, but a companion $\textnormal{head}_j^{z,\tau}$ with an endorsement level of $\tau$ attends only to source tokens that are endorsed $\tau$ times or more.
This is achieved with a special binary mask $M_{i}^{\tau}$ (Eqs.~(\ref{eq:head}-\ref{eq:mask_endorse})).
The original heads are believed to copy over source tokens that are deemed relevant to summary tokens according to the dependency syntax~\cite{clark-etal-2019-bert}.
The companion heads serve a similar purpose but have a narrower focus on endorsed source tokens---frequently endorsed tokens are more likely to be copied over by companion heads.
The method thus improves head diversity similar to that of sparse Transformers~\cite{correia-etal-2019-adaptively,huang-etal-2021-efficient}.  
The hyperparameter $\tau$ controls the level of endorsement.
Finally, all heads are pooled into a hidden vector $\widehat{\bm{g}}_j^{(l)}$ (Eq.~(\ref{eq:combine})) to be passed to the feedforward layer.
\begin{align*}
& \bm{q}_j^z = \bm{W}_z^Q \textcolor{black}{\widetilde{\bm{g}}_j^{(l-1)}} \,\;\;\quad j \in [n]\numberthis\label{eq:query}\\
& \bm{k}_i^z = \bm{W}_z^K \textcolor{black}{\bm{h}_i^{(L)}} \quad\quad i \in [m]\numberthis\label{eq:key}\\
& \bm{v}_i^z = \bm{W}_z^V \textcolor{black}{\bm{h}_i^{(L)}} \quad\quad i \in [m]\numberthis\label{eq:value}\\
& \textcolor{black}{\textnormal{head}_j^{z,\tau}} = \sum_{i=0}^m \frac{\exp(\bm{q}_j^{z\top} \bm{k}_i^z)}{\sum_{r=0}^m \exp(\bm{q}_j^{z\top} \bm{k}_{r}^z)} \textcolor{black}{M_{i}^{\tau}} \bm{v}_i^z \numberthis\label{eq:head}\\
&M_{i}^{\tau} = \left\{
\begin{array}{ll}
1 & \text{if} \ \textnormal{Endorse}(x_i) \geq \tau\\
0 & \text{otherwise}
\end{array}
\right.\numberthis\label{eq:mask_endorse}\\
& \widehat{\bm{g}}_j^{(l)} = \sum_{z=1}^{n_{\textnormal{head}}} \sum_{\tau=0}^{\tau_{\textnormal{max}}} \textcolor{black}{\textnormal{head}_j^{z,\tau}}\bm{W}_z^{\tau} \numberthis\label{eq:combine}
\end{align*}

When $\tau_{\textnormal{max}}$ is set to 0, the model reduces to its initial form using the original heads, i.e., $\textnormal{head}_j^{z,0}$.
Further, we initialize
$\bm{W}_z^{\tau} = \lambda^{\tau} \bm{W}_z$, where $\bm{W}_z \in \mathbb{R}^{h_{\textnormal{head}} \times h_{\textnormal{model}}}$ are pretrained model parameters associated with the head $z$.
$\lambda^{\tau} \in [0,1]$ is a coefficient and $\bm{W}_z = \sum_{\tau=0}^{\tau_{\textnormal{max}}} \bm{W}_z^{\tau}$. 
It indicates that, head $z$ and all of its companion heads are linearly interpolated to produce decoder hidden state $\widehat{\bm{g}}_j^{(l)}$. 
If a source token is not endorsed, it will have a reduced impact on the decoder hidden state when companion heads are used.
The method has the advantage that,
when new documents are dynamically added or removed from the set, it only changes the level of endorsement received by the tokens ($\tau$), thus avoiding costly retraining of the neural encoder-decoder model.
We proceed by describing how fine-grained segment-level endorsement is obtained from modeling synopsis-document relationships.

\section{Modelling Endorsement}
\label{sec:endorsement}

In this section, we present the first stage in our approach -- modelling endorsement -- whose outputs are passed to the abstractive summarization model in the second stage.
Modelling endorsement serves two main purposes.
It allows us to identify salient segments of text using a frequency-driven endorsement model, and the level of endorsement guides the summarizer to consolidate salient content.
Further, it helps us reduce the source input from multiple documents to a single pseudo-document, whereby any unendorsed sentences are removed from consideration.

A fragment of text is considered to be endorsed if its information is observed in the endorser.
We obtain a set of synopses from the source documents; they are used as \emph{endorsers} to identify salient segments from a candidate source document.
A segment that is endorsed only once indicates its information is considered important by only one source document. 
Frequent endorsement by multiple endorsers suggests the information is reiterated in multiple source documents, and reiteration implies increased salience.
Any information that is present among multiple sources is likely to be important.
Thus, our method identifies salient segments considering both within- and cross-document saliency.
Our approach is in spirit similar to those of building semantic concept graphs for multi-document summarization \cite{bing-etal-2015-abstractive,handler-oconnor-2018-relational,falke-gurevych-2019-fast} in that frequently reiterated concepts are likely to be captured.
However, we do not explicitly construct semantic concept graphs, but focus on modeling synopsis-document endorsement and incorporating it into summary generation, which distinguishes our work from these studies.
We investigate two variants to compute segment-level endorsement.

\subsection{Synopsis-Document Alignment}
\label{sec:alignment}

Let $S$ be a synopsis serving as the endorser and $D$ a source document, our goal is to estimate whether a token $x_i$ of the document is endorsed by the synopsis.
A soft alignment between the synopsis and document is attainable by utilizing text evaluation metrics such as BERTScore~\cite{Zhang2020BERTScore}, 
where we build contextualized embeddings for tokens of the document and synopsis, compute the cosine similarity of embeddings, and find a most similar synopsis token for each token of the document to obtain the endorsement score $\mathcal{S}(x_i)$ (Eq.~(\ref{eq:sim})).
Albeit a greedy alignment, the method can produce competitive results comparing to methods such as the earth mover's distance~\cite{zhao-etal-2019-moverscore}.
\begin{align*}
\mathcal{S}(x_i) = \max_{y_j \in S} \textnormal{Sim}(x_i, y_j) \numberthis\label{eq:sim}
\end{align*}

\noindent\textbf{Contiguous Segments}\quad 
It is important to endorse segments of text rather than isolated tokens, as segments such as ``\emph{\$8 million}'' is either included in the abstract in its entirety, or not at all.
We transform token-level endorsement scores into binary decisions using the maximum sum subarray algorithm (Eq.~(\ref{eq:mcss})), which finds a contiguous subsequence that yields the highest sum of scores. 
The solution is trivial when all scores are positive. 
We thus offset the scores by $\delta$ before applying the algorithm.
Let $\{0.2, 0.3, -0.1, 0.4, -0.5\}$ be an example of a set of adjusted endorsement scores, the algorithm endorses the first four tokens as the sum of their scores is the highest, yielding $\{1, 1, 1, 1, 0\}$, where 1 indicates the token is endorsed and 0 otherwise.
We apply the algorithm to each sentence of the document and discard the segment if it has less than 5 tokens.
The method endorses salient segments of text, yet is lenient to include gap tokens.
\begin{align*}
\{s, e\} = \argmax_{\{i,j\} \in m} \sum_{k=i}^j (\mathcal{S}(x_k)-\delta) \numberthis\label{eq:mcss}
\end{align*}

\noindent\textbf{Soft vs. Hard Alignment}\quad 
A hard alignment between the synopsis and document can be obtained from string matching.
A document token receives a score of 1 if it finds a match in the synopsis.
Similar to above, we offset the scores by $\delta$ to obtain segments of endorsed text.
Hard alignment is sensitive to entities and quantities; yet it can miss out on paraphrases. 
We compare the effectiveness of these alignment methods in the results section.

\subsection{Synopses as Endorsers}
\label{sec:synopses_as_endorsers}

A synopsis contains the main points of the source document.
We employ BART~\cite{lewis-etal-2020-bart}, fine-tuned on the CNN/DailyMail dataset, as a single-document abstractive summarizer to produce a synopsis from each document of the input cluster. 
Synopses as endorsers are superior to whole documents or sentence extracts.
Not only are synopses more concise, but they can exclude superfluous information such as quoted material from consideration.
We score all sentences of the source documents according to the sum of their token endorsement scores.
Highest endorsed sentences are selected and arranged in chronological order to form a pseudo-document, with a limit of $|D|$ tokens, which serves as the input to our summarization module.

When a token is deemed salient by $\tau$ endorsers, we set $\textnormal{Endorse}(x_i)$=$\tau$, analogous to a majority vote by the pool of endorsers.
We introduce two endorsement patterns. \emph{Reciprocal endorsement} is where a synopsis can endorse every document of the cluster, akin to how every token attends to every other token in Transformer self-attention. \emph{Sequential endorsement} is where source documents are arranged in chronological order and only synopses of the later documents can endorse the earlier documents, akin to how each token can attend only to previous tokens in decoder-only self-attention.
Sequential endorsement assumes the first few articles of an event or topic are more important than others.
It avoids endorsing redundant content, which is particularly useful when the documents contain redundancy or noise that is typical in the output of clustering algorithms for content aggregation. 
Importantly, our endorsement framework offers a potential to customize endorsement patterns based on the trustworthiness of news sources, political leanings, content quality, and more.

\section{Data}
\label{sec:data}

We experiment with a large-scale multi-document summarization dataset~\cite{gholipour-ghalandari-etal-2020-large} whose data are gathered from the Wikipedia Current Events Portal (WCEP).\footnote{\url{https://en.wikipedia.org/wiki/Portal:Current_events}}
The dataset contains an archive of important news events happening around 2016--2019.
Each event is associated with a succinct summary of 30-40 words written by the editor and an average of 1.2 source articles linked from the event page. 
Additional source articles are retrieved from the CommonCrawl-News dataset using an event classifier.
These articles are published within a window of $\pm$1 day of the event date. 
We sample from these additional articles to ensure each event has 10 source articles.
All summaries and source articles are in English.
The dataset contains 8,158, 1,020 and 1,022 clusters respectively in the train, validation and test splits.

Our method aims to produce an abstractive summary from a cluster of news articles discussing a given event or topic.
To assess the generality of our method, 
we apply the model trained on WCEP to three different test sets, i.e., the test split of WCEP and two benchmark multi-document summarization datasets, DUC-04 and TAC-11.
The DUC/TAC datasets contain 50 and 44 clusters, respectively.
They each comprise a set of news events collected over a period of time, and thus are suitable for evaluation of the model's generality in out-of-domain scenarios.
DUC and TAC datasets contain four reference summaries per cluster created by NIST evaluators. WCEP has a single reference summary per cluster written by editors.
The target summary length is 100 words for DUC/TAC and 40 words for WCEP, following the convention of previously published results. 
Endorsement-related statistics for these datasets are presented in Table~\ref{tab:tau_values}.

\section{Experiments}

\vspace{0.05in}
\emph{\textbf{Baseline Systems.}}\quad
We compare our endorsement method to strong multi-document summarization baselines.
The extractive summarization systems include
(i) \textit{TextRank}~\cite{mihalcea-tarau-2004-textrank} and \textit{LexRank}~\cite{Erkan:2004}, which are graph-based approaches that perform strongly on this task.
(ii) \textit{Centroid}~\cite{Hong:2014} computes the importance of a source sentence based on its cosine similarity with the document centroid.
(iii) \textit{Submodular}~\cite{lin-bilmes-2011-class} treats multi-document summarization as a submodular maximization problem.
(iv) \textit{KL-Sum}~\cite{haghighi-vanderwende-2009-exploring} is a greedy approach that adds sentences to the summary to minimize KL divergence.
(v) \textit{TSR} and \textit{BertReg}~\cite{gholipour-ghalandari-etal-2020-large} are regression-based sentence ranking methods using averaged word embeddings (TSR) and BERT sentence embeddings (BertReg).

The abstractive summarization systems include:
(vi) \textit{PointerGen}~\cite{see-etal-2017-get}, which generates a summary by copying source words and predicting new words.
The set of documents are concatenated to form the input.
(vii) \textit{PG-MMR}~\cite{lebanoff-etal-2018-adapting} exploits the maximal marginal relevance method to select sentences and an encoder-decoder model to fuse them into an abstract.
(viii) \textit{Hi-MAP}~\cite{fabbri-etal-2019-multi} introduces an end-to-end hierarchical attention model to generate abstracts from multi-document inputs.
We compare our system to these baselines and report results on WCEP, DUC-04, and TAC-11 datasets\footnote{
We were unable to compare our method with hierarchical Transformers \cite{liu-lapata-2019-hierarchical} because the authors did not make their ranker available for ranking paragraphs. 
}.

\begin{table}[t]
\centering
\setlength{\tabcolsep}{4pt}
\renewcommand{\arraystretch}{1.15}
\begin{footnotesize}
\begin{tabular}{|l|ccc||rrr|}
\hline
\rowcolor{gray!10}
& \textbf{Synop} & \textbf{Num} & \textbf{Seg} & \multicolumn{3}{c|}{\% \textbf{Endorse Scores} $\geq \tau$} \\
\rowcolor{gray!10}
\textbf{Dataset}  & \textbf{Len} & \textbf{Segs} & \textbf{Len} & $\tau=0$ & $\tau=1$ & $\tau=2$\\
\hline
WCEP & 61 & 4.9 & 14.2 & 100.0 & 12.6 & 5.6 \\
DUC-04 & 58 & 6.1 & 11.7 & 100.0 & 9.7 & 2.3 \\
TAC-11 & 60 & 6.7 & 11.8 & 100.0 & 14.5 & 4.1 \\
\hline
\end{tabular}
\end{footnotesize}
\vspace{-0.05in}
\caption{
(\textsc{Left}) The average length of synopses (SynopLen), average number of segments in a source document endorsed by a synopsis and average length of endorsed segments (SegLen).
(\textsc{Right}) Percentage of tokens with endorsement scores above the threshold value used in each set of companion heads. All tokens with scores below the threshold are masked out.}
\vspace{-0.2in}
\label{tab:tau_values}
\end{table}

\vspace{0.05in}
\emph{\textbf{Sequential vs. Reciprocal Endorsement.}}\quad
We investigate two endorsement patterns: (a) \emph{reciprocal endorsement} allows any two documents of the same cluster to endorse each other, and (b) \emph{sequential endorsement} arranges source documents in chronological order and only later documents are allowed to endorse earlier ones.
The endorsement mechanism provides the flexibility needed for many domains to exploit cross-document relationships to generate abstractive summaries. 
For our variants, the highest-scoring sentences are consolidated to form an input document which, along with the endorsement scores, are passed to our endorsement-aware abstractor to be condensed into a summary.

\vspace{0.05in}
\emph{\textbf{Endorsement-Aware Abstractor.}}\quad
We employ BART, a state-of-the-art encoder-decoder model as our base abstractor~\cite{lewis-etal-2020-bart}. 
The model has 24 layers in the encoder and decoder, a hidden size of 1024, 16 heads, with a total of 406M parameters.
It was fine-tuned on the train split of WCEP for an average of two epochs with a batch size of 4. 
We use the Adam optimizer~\cite{Kingma:2015} and a learning rate of $3^{-5}$ with warm-up.
At inference time, we use a beam size of $K$=4, with a minimum decoding length of 10 and a maximum of 50 tokens. 
Our implementation is based on fairseq\footnote{\url{https://github.com/pytorch/fairseq}} and it takes about two hours to train the model on a NVIDIA V100 32GB GPU card.

For the endorsement-aware abstractor, we add two sets of companion heads to the decoder, for a total of 48 attention heads. The $\tau$ values for each set of heads are 0/1/2. 
Table~\ref{tab:tau_values} shows the percentage of tokens that receive different levels of attention: 12\% of the tokens receive level-$1$ attention ($\tau=1$), 4\% receive level-$2$ attention ($\tau=2$). 
The $\lambda^{\tau}$ values are set to be 0.8, 0.1, and 0.1---this gives more influence to the original attention heads, so the model is not confused by the addition of the new heads that attend to endorsed segments.
We use a maximum of 1024 tokens for the input document.

\vspace{0.05in}
\emph{\textbf{Synopsis-Document Endorsement.}}\quad
To enable soft alignment between a synopsis and a candidate document, we use BERTScore~\cite{Zhang2020BERTScore} with the following hash code: \textsf{\footnotesize roberta-large\_L17\_no-idf\_version=0.3.2(hug\_trans=2.8.0)-rescaled}. 
It suggests that the token representations are drawn from the 17th layer of RoBERTa-large.
Our maximum sum subarray algorithm requires the scores to contain a mix of positive/negative values. 
Thus, we subtract all scores by $\delta$. The $\delta$ values are 0.85 and 0.8 for the soft and hard alignment, respectively. 
These values are tuned on validation data, where a larger $\delta$ indicates fewer tokens will be endorsed.

We proceed by presenting summarization results on our datasets, including an ablation study to examine the contribution of each part of our method. We also present a case study showcasing the potential of our endorsement method.

\begin{table}[t]
\setlength{\tabcolsep}{7pt}
\renewcommand{\arraystretch}{1.15}
\centering
\begin{footnotesize}
\textsf{
\begin{tabular}{|lrrr|}
\hline
\textbf{System} & \textbf{R-1} & \,\,\,\textbf{R-2} & \textbf{R-SU4} \\
\hline
\rowcolor{gray!10}
\multicolumn{4}{|l|}{\textbf{Extractive}}\\
Random Lead\quad\quad\quad & 27.6 & 9.1 & --\\
Random & 18.1 & 3.0 & --\\
TextRank & 34.1 & 13.1 & --\\
Centroid & 34.1 & 13.3 & --\\
Submodular & 34.4 & 13.1 & --\\
TSR & 35.3 & 13.7 & --\\
BertReg & 35.0 & 13.5 & --\\
\hline
\rowcolor{gray!10}
\multicolumn{4}{|l|}{\textbf{Our Method (In-Domain)}}\\
Endorser-Reciprocal & 43.3 & 21.9 & 22.1 \\
Endorser-Sequential & \textbf{45.4} & \textbf{23.2} & \textbf{23.5} \\
\hline
\end{tabular}}
\end{footnotesize}
\vspace{-0.05in}
\caption[Caption for LOF]{
A comparison of multi-document summarizers on WCEP's test set.\protect\footnotemark
\emph{Endorser-*} are our proposed methods.
}
\label{tab:results_wcep}
\vspace{-0.15in}
\end{table}
\footnotetext{
We note that baseline summarizers use a maximum of 100 articles per cluster;
these results are obtained from Gholipour Ghalandari et al.~\shortcite{gholipour-ghalandari-etal-2020-large}.
In contrast, our endorsement methods outperform the baselines with only 10 input articles per cluster. 
}

\subsection{Results}
\label{sec:results}

Our methods achieve state-of-the-art results when compared to previous work on WCEP's test set (Table~\ref{tab:results_wcep}). Sequential endorsement outperforms reciprocal endorsement due to the ability of sequential endorsement to remove redundancies introduced in later documents.
In news domain, later articles generally review information from previous articles and introduce small developments in the story. By ordering the documents chronologically and having later articles give endorsement to earlier articles, it encourages the summarizer to pick content from earlier articles and reduce redundancy introduced in later articles.
The largest performance increase can be seen in R-2, with \emph{Endorser-Sequential} achieving a 9.7 increase over a BERT-based method. It demonstrates the effectiveness of endorsement for detecting salient segments and stitching them together to form a summary.

\begin{table}[t]
\setlength{\tabcolsep}{7pt}
\renewcommand{\arraystretch}{1.15}
\centering
\begin{footnotesize}
\textsf{
\begin{tabular}{|lrrr|}
\hline
\textbf{System} & \textbf{R-1} & \,\,\,\textbf{R-2} & \textbf{R-SU4} \\
\hline
\rowcolor{gray!10}
\multicolumn{4}{|l|}{\textbf{Extractive}}\\
TextRank & 33.16 & 6.13 & 10.16 \\
LexRank & 34.44 & 7.11 & 11.19 \\
Centroid & 35.49 & 7.80 & 12.02 \\
\hline
\rowcolor{gray!10}
\multicolumn{4}{|l|}{\textbf{Neural Abstractive}}\\
Pointer-Gen & 31.43 & 6.03 & 10.01\\
PG-MMR & {36.88} & 8.73 & {12.64}\\
PG-BRNN & 29.47 & 6.77 & 7.56\\
Hi-MAP & 35.78 & {8.90} & 11.43\\
\hline
\rowcolor{gray!10}
\multicolumn{4}{|l|}{\textbf{Our Method (Out-of-Domain)}}\\
Endorser-Sequential & 34.74 & 8.08 & 12.06 \\
Endorser-Reciprocal & {35.24} & {8.61} & {12.49} \\
Endorser-Oracle & 36.27 & {8.93} & {13.04}\\
\hline
\end{tabular}}
\end{footnotesize}
\vspace{-0.05in}
\caption{A comparison of multi-document summarizers on the DUC-04 dataset.
\emph{Endorser-*} are our methods.
}
\label{tab:results_duc04}
\vspace{-0.1in}
\end{table}

\begin{table}[t]
\setlength{\tabcolsep}{6.5pt}
\renewcommand{\arraystretch}{1.15}
\centering
\begin{footnotesize}
\textsf{
\begin{tabular}{|lrrr|}
\hline
\textbf{System} & \textbf{R-1} & \,\,\,\textbf{R-2} & \textbf{R-SU4} \\
\hline
\hline
Endorser-HardAlign & 44.7 & 22.4 & 22.6 \\
Endorser-SoftAlign & 45.4 & 23.2 & 23.5 \\
\:\:\:\:- companion heads & \textbf{45.8} & \textbf{23.5} & \textbf{23.8} \\
\:\:\:\:- endorse selection & 43.6 & 23.0 & 22.9 \\
\:\:\:\:- abstractive module & 28.3 & 9.3 & 10.9 \\
\hline
\end{tabular}}
\end{footnotesize}
\vspace{-0.05in}
\caption{Ablation study on WCEP dataset.}
\vspace{-0.15in}
\label{tab:ablation}
\end{table}

We report experimental results on DUC-04 and TAC-11 datasets in Tables~\ref{tab:results_duc04} and \ref{tab:results_tac11}.
Here, our methods can outperform or perform comparably to previous summarization methods.
On the WCEP test set, it corresponds to an \textbf{\emph{in-domain}} scenario.
On DUC-04 and TAC-11 test sets, it is an \textbf{\emph{out-of-domain}} scenario.
Due to data scarcity, the model can only be trained on the train split of WCEP and then tested on DUC/TAC datasets. 
The fact that our system, when used out-of-the-box, can attain better or comparable results to the previous state-of-the-art has demonstrated its strong generalization capability.
It suggests that obtaining segment-level endorsement on an outside domain then using it to inform summary generation is meaningful.

We observe that the reciprocal endorsement strategy outperforms sequential endorsement for DUC-04 and TAC-11 test sets. A closer look at the data suggests that this is due to the lower amount of redundancy present in DUC/TAC data. 
While WCEP documents are \emph{automatically clustered} and contain much redundancy, source documents of DUC/TAC are \emph{manually selected} by NIST assessors, each successive document in a topic cluster presents new developments about the topic. Thus, reciprocal endorsement may lead to better results for domains with less redundancy.

Intuitively, we want to steer the model attention towards endorsed segments if they are of high quality, and away from the segments otherwise.
We conduct a set of oracle experiments that set $\lambda^{\tau}$ values to be proportional to the R-2 recall scores of endorsed segments (\emph{Endorser-Oracle}).
If the segments obtained for $\tau=2$ yield a high R-2 recall score, they contain summary content and the model should attend to these endorsed segments by using a high $\lambda^{\tau}$ value.
Results are reported in Tables~\ref{tab:results_duc04} and \ref{tab:results_tac11}. 
We find that such a strategy is effective for making the most of companion heads. 
Future work may associate attention ($\lambda^{\tau}$ values) with the quality of segments obtained at different levels of endorsement ($\tau=\{0,1,2\}$).

\begin{table}[t]
\setlength{\tabcolsep}{7pt}
\renewcommand{\arraystretch}{1.15}
\centering
\begin{footnotesize}
\textsf{
\begin{tabular}{|lrrr|}
\hline
\textbf{System} & \textbf{R-1} & \,\,\,\textbf{R-2} & \textbf{R-SU4} \\
\hline
\rowcolor{gray!10}
\multicolumn{4}{|l|}{\textbf{Extractive}}\\
KLSumm & 31.23 & 7.07 & 10.56 \\
LexRank & 33.10 & 7.50 & 11.13 \\
\hline
\rowcolor{gray!10}
\multicolumn{4}{|l|}{\textbf{Neural Abstractive}}\\
Pointer-Gen & 31.44 & 6.40 & 10.20\\
PG-MMR & 37.17 & {10.92} & {14.04}\\
\hline
\rowcolor{gray!10}
\multicolumn{4}{|l|}{\textbf{Our Method (Out-of-Domain)}}\\
Endorser-Sequential & 36.11 & 9.52 & 13.07 \\
Endorser-Reciprocal & {37.43} & {10.71} & {13.94} \\
Endorser-Oracle & {38.01} & {11.11} & {14.61}\\
\hline
\end{tabular}}
\end{footnotesize}
\vspace{-0.05in}
\caption{A comparison of multi-document summarizers on the TAC-11 test set.
\emph{Endorser-*} are our methods.
}
\vspace{-0.15in}
\label{tab:results_tac11}
\end{table}

\begin{table*}
\setlength{\tabcolsep}{0pt}
\renewcommand{\arraystretch}{1}
\centering
\begin{scriptsize}
\begin{tabular}{l}
\toprule
\textbf{\color{red}{(a) Single Synopsis Generated by BART}}\\[0.2em]
\textsf{Opposition leader Sam Rainsy seeks clarification of security guarantees promised by Hun Sen. Hun Sen announced a government guarantee of}\\
\textsf{all politicians' safety Wednesday. The opposition leader was forced to take refuge in a U.N. office in September to avoid arrest. The two parties}\\
\textsf{have formed three working groups to hammer out details of the agreement.}\\
\midrule
\textbf{\color{red}{(b) Endorsement from All Synopses}}\\[0.2em]
\textsf{\colorbox{orange!10}{Sam Rainsy,} who earlier called Hun Sen's statement "full of loopholes," asked Sihanouk for his help in obtaining \colorbox{orange!10}{a promise from Hun Sen that all}}\\ 
\textsf{members of the Sam Rainsy Party were free from prosecution for their political activities during and after last July's election. \colorbox{orange!10}{Sam Rainsy, a staunch}}\\ 
\textsf{critic of Hun Sen, \colorbox{orange!20}{was forced to take refuge in a U.N. office in September to avoid arrest} after Hun Sen accused him of being behind a plot against}\\ 
\textsf{his life. The alleged assassination attempt came during massive street demonstrations organized by \colorbox{orange!30}{ the opposition after Hun Sen's Cambodian}}\\ 
\textsf{People's Party narrowly won the election. The opposition, alleging widespread fraud and intimidation, refused to accept the results of the polls.}\\
\textsf{Fearing for their \colorbox{orange!10}{safety, Sam Rainsy} and his then-ally \colorbox{orange!40}{Prince Norodom Ranariddh} led an exodus of opposition lawmakers out of Cambodia}\\ 
\textsf{after parliament was ceremonially opened in late September. Ranariddh, whose \colorbox{orange!10}{FUNCINPEC party} finished a close second in the election,}\\ 
\textsf{returned \colorbox{orange!10}{last week and struck a deal with}\colorbox{orange!20}{ Hun Sen}\colorbox{orange!40}{ to form a coalition government.} The agreement will make Hun Sen prime minister and}\\
\textsf{Ranariddh president \colorbox{orange!20}{of the National Assembly. The two parties have formed three working groups to} \colorbox{orange!10}{hammer out details of} \colorbox{orange!20}{the agreement,}}\\
\textsf{\colorbox{orange!20}{including the establishment of a Senate to}\colorbox{orange!10}{ be the upper house of parliament.} Sok An, representing\colorbox{orange!30}{ Hun Sen's party}, said...}\\
\midrule
\textbf{\color{red}{(c) Human-Chosen Segments}}\\[0.2em]
\textsf{Sam Rainsy, who earlier called Hun Sen's statement "full of loopholes," asked Sihanouk for his help in obtaining a promise from Hun Sen that all}\\
\textsf{members of the Sam Rainsy Party were free from prosecution for their political activities during and after last July's election. Sam Rainsy, a staunch}\\
\textsf{critic of Hun Sen, was \colorbox{orange!40}{forced to take refuge} in a U.N. office in September \colorbox{orange!40}{to avoid arrest}\colorbox{orange!10}{ after} \colorbox{orange!10}{Hun Sen accused him of being behind a plot against}}\\
\textsf{\colorbox{orange!10}{his life. The alleged assassination attempt came during massive street}\colorbox{orange!10}{ demonstrations} organized by the opposition after \colorbox{orange!10}{Hun Sen's Cambodian}}\\
\textsf{\colorbox{orange!10}{People's Party narrowly won the election. The opposition, alleging widespread fraud and intimidation,} refused to accept the results of the polls.}\\
\textsf{\colorbox{orange!40}{Fearing for their safety, Sam Rainsy and his then-ally Prince Norodom Ranariddh led an exodus of opposition lawmakers out of Cambodia}}\\
\textsf{after parliament was ceremonially opened in late September. Ranariddh, whose FUNCINPEC party finished a close second in the election,}\\
\textsf{returned last week and\colorbox{orange!30}{ struck a deal with Hun Sen to form a coalition government.} \colorbox{orange!30}{The agreement will make Hun Sen prime minister and}}\\
\textsf{\colorbox{orange!30}{Ranariddh president of the National Assembly.} The two parties have formed three working groups to hammer out details of the agreement,}\\
\textsf{including the establishment of a Senate to be the upper house of parliament. Sok An, representing Hun Sen's party, said...}\\
\bottomrule
\end{tabular}
\end{scriptsize}
\vspace{-0.1in}
\caption{An analysis of endorsed segments for a document. 
\textbf{(a)} A synopsis is generated from a candidate document.
\textbf{(b)} The document also receives endorsement from the other 9 synopses in the cluster. 
\textbf{(c)} We compare to segments chosen by a human using the Pyramid method. 
Stronger highlighting indicates the segment received endorsement from many synopses.}
\label{tab:case_study}
\vspace{-0.1in}
\end{table*}

\subsection{Ablation}

We perform an ablation study on WCEP to study the effects of each component in our model (Table \ref{tab:ablation}). 
First, we compare the endorsement methods, denoted by \textit{HardAlign} and \textit{SoftAlign}. 
SoftAlign achieves consistently better results, showing that it is important to allow flexibility when aligning synopses to documents for endorsement. 
Next, we remove several components from the best-performing model (SoftAlign) to understand the effect of each. Removing ``companion heads'' from the abstractive model results in a very small boost in performance. Removing ``endorsement selection''---meaning the model uses no information gained from performing endorsement, and is simply a BART model trained to summarize documents---leads to a significant performance drop, especially in R-1. It suggests that using endorsement to identify summary-worthy content from multiple documents is beneficial for an abstractive model.

Moreover, removing the ``abstractive model''---meaning summaries are created extractively by selecting the highest-endorsed sentences---results in a large decrease in scores. It indicates that content-selection by endorsement cannot be done alone without an abstractor to create a more concise summary.
This is especially the case for WCEP, where human reference summaries are relatively short.

We additionally report BERTScore~\cite{Zhang2020BERTScore} to evaluate summaries, in addition to the ROUGE metric~\cite{lin-2004-rouge}. 
BERTScore uses cosine similarity between BERT contextual embeddings of words to detect word overlap between two texts, thus overcoming the problem of lexical variation in summarization. 
On DUC-04, the $F_1$ scores are 29.89 and 30.14, respectively for our sequential and reciprocal model.
The score for human reference summary is 35.08.
They show very similar trends to those in Table~\ref{tab:results_duc04}, suggesting that our method when tested in out-of-domain scenarios can achieve competitive results.

\subsection{A Case Study}
\label{sec:case_study}

We present an in-depth analysis of our fine-grained endorsement in Table~\ref{tab:case_study}. 
Soft alignment is used to endorse a candidate document from synopses of the cluster. 
We compare the resulting endorsements to the text segments chosen by a human using the Pyramid method~\cite{nenkova-passonneau-2004-evaluating}, where semantic content units (SCUs) are identified from the reference summaries and are matched to phrases in the candidate document. 
The segments selected by our endorsement method and those chosen by manual annotation show a great amount of overlap, exemplifying the strength of our method in locating salient content from multi-document inputs. In fact, our endorsement method draws strong parallels with the Pyramid method---in our case, sentences from the automatically-generated synopses act as SCUs, which are then matched to phrases in the candidate document using a soft or hard alignment.

We observe that the endorsement given by a single synopsis is already quite similar to the human segments. However, taking the average endorsement from all ten synopses results in a higher quality set of segments. 
This shows the inherent value that exists from repetition in multi-document clusters, and it shows the importance of leveraging all of the documents rather than just a single one for salience estimation.
Importantly, we observe that named entities, e.g., \textit{``Sam Rainsy,''} \textit{``King Norodom Sihanouk,''} are more readily endorsed than other phrases. These entities are frequently repeated verbatim in all of the documents, thereby increasing their likelihood of being endorsed. 

We envision future neural document summarization systems to produce better synopses than BART. 
They can lead to more accurate estimates for endorsed segments, hence improving the overall performance of our multi-document summarizer. 
The endorsement mechanism at its core is simple and robust---looking for shared content between a document and a synopsis.
It provides great flexibility allowing the summarizer to potentially operate on document clusters containing a varying number of documents, 
which is a desirable characteristic.

\section{Conclusion}
\label{sec:conclusion}

We presented a novel framework to model asynchronous endorsement between synopses and documents for multi-document abstractive summarization.
We introduced an endorsement method to enrich the encoder-decoder model with fine-grained endorsement.
Our method was evaluated on benchmark multi-document datasets and we discussed challenges and shed light on future research.

\section*{Acknowledgments}

We are grateful to the anonymous reviewers for their helpful comments and suggestions.
This research was supported in part by the National Science Foundation grant IIS-1909603.

\bibliographystyle{acl_natbib}
\bibliography{anthology,more,summ,logan}

\begin{thebibliography}{53}
\expandafter\ifx\csname natexlab\endcsname\relax\def\natexlab#1{#1}\fi

\bibitem[{Berg-Kirkpatrick et~al.(2011)Berg-Kirkpatrick, Gillick, and
  Klein}]{berg-kirkpatrick-etal-2011-jointly}
Taylor Berg-Kirkpatrick, Dan Gillick, and Dan Klein. 2011.
\newblock \href {https://aclanthology.org/P11-1049} {Jointly learning to
  extract and compress}.
\newblock In \emph{Proceedings of the 49th Annual Meeting of the Association
  for Computational Linguistics: Human Language Technologies}, pages 481--490,
  Portland, Oregon, USA. Association for Computational Linguistics.

\bibitem[{Bing et~al.(2015)Bing, Li, Liao, Lam, Guo, and
  Passonneau}]{bing-etal-2015-abstractive}
Lidong Bing, Piji Li, Yi~Liao, Wai Lam, Weiwei Guo, and Rebecca Passonneau.
  2015.
\newblock \href {https://doi.org/10.3115/v1/P15-1153} {Abstractive
  multi-document summarization via phrase selection and merging}.
\newblock In \emph{Proceedings of the 53rd Annual Meeting of the Association
  for Computational Linguistics and the 7th International Joint Conference on
  Natural Language Processing (Volume 1: Long Papers)}, pages 1587--1597,
  Beijing, China. Association for Computational Linguistics.

\bibitem[{Boudin et~al.(2015)Boudin, Mougard, and
  Favre}]{boudin-etal-2015-concept}
Florian Boudin, Hugo Mougard, and Benoit Favre. 2015.
\newblock \href {https://doi.org/10.18653/v1/D15-1220} {Concept-based
  summarization using integer linear programming: From concept pruning to
  multiple optimal solutions}.
\newblock In \emph{Proceedings of the 2015 Conference on Empirical Methods in
  Natural Language Processing}, pages 1914--1918, Lisbon, Portugal. Association
  for Computational Linguistics.

\bibitem[{Bra{\v{z}}inskas et~al.(2020)Bra{\v{z}}inskas, Lapata, and
  Titov}]{brazinskas-etal-2020-shot}
Arthur Bra{\v{z}}inskas, Mirella Lapata, and Ivan Titov. 2020.
\newblock \href {https://doi.org/10.18653/v1/2020.emnlp-main.337} {Few-shot
  learning for opinion summarization}.
\newblock In \emph{Proceedings of the 2020 Conference on Empirical Methods in
  Natural Language Processing (EMNLP)}, pages 4119--4135, Online. Association
  for Computational Linguistics.

\bibitem[{Brown et~al.(2020)Brown, Mann, Ryder, Subbiah, Kaplan, Dhariwal,
  Neelakantan, Shyam, Sastry, Askell, Agarwal, Herbert-Voss, Krueger, Henighan,
  Child, Ramesh, Ziegler, Wu, Winter, Hesse, Chen, Sigler, Litwin, Gray, Chess,
  Clark, Berner, McCandlish, Radford, Sutskever, and
  Amodei}]{brown2020language}
Tom~B. Brown, Benjamin Mann, Nick Ryder, Melanie Subbiah, Jared Kaplan,
  Prafulla Dhariwal, Arvind Neelakantan, Pranav Shyam, Girish Sastry, Amanda
  Askell, Sandhini Agarwal, Ariel Herbert-Voss, Gretchen Krueger, Tom Henighan,
  Rewon Child, Aditya Ramesh, Daniel~M. Ziegler, Jeffrey Wu, Clemens Winter,
  Christopher Hesse, Mark Chen, Eric Sigler, Mateusz Litwin, Scott Gray,
  Benjamin Chess, Jack Clark, Christopher Berner, Sam McCandlish, Alec Radford,
  Ilya Sutskever, and Dario Amodei. 2020.
\newblock \href {http://arxiv.org/abs/2005.14165} {Language models are few-shot
  learners}.

\bibitem[{Chen and Bansal(2018)}]{chen-bansal-2018-fast}
Yen-Chun Chen and Mohit Bansal. 2018.
\newblock \href {https://doi.org/10.18653/v1/P18-1063} {Fast abstractive
  summarization with reinforce-selected sentence rewriting}.
\newblock In \emph{Proceedings of the 56th Annual Meeting of the Association
  for Computational Linguistics (Volume 1: Long Papers)}, pages 675--686,
  Melbourne, Australia. Association for Computational Linguistics.

\bibitem[{Cho et~al.(2019{\natexlab{a}})Cho, Lebanoff, Foroosh, and
  Liu}]{cho-etal-2019-improving}
Sangwoo Cho, Logan Lebanoff, Hassan Foroosh, and Fei Liu. 2019{\natexlab{a}}.
\newblock \href {https://doi.org/10.18653/v1/P19-1098} {Improving the
  similarity measure of determinantal point processes for extractive
  multi-document summarization}.
\newblock In \emph{Proceedings of the 57th Annual Meeting of the Association
  for Computational Linguistics}, pages 1027--1038, Florence, Italy.
  Association for Computational Linguistics.

\bibitem[{Cho et~al.(2019{\natexlab{b}})Cho, Li, Yu, Foroosh, and
  Liu}]{cho-etal-2019-multi}
Sangwoo Cho, Chen Li, Dong Yu, Hassan Foroosh, and Fei Liu. 2019{\natexlab{b}}.
\newblock \href {https://doi.org/10.18653/v1/D19-5412} {Multi-document
  summarization with determinantal point processes and contextualized
  representations}.
\newblock In \emph{Proceedings of the 2nd Workshop on New Frontiers in
  Summarization}, pages 98--103, Hong Kong, China. Association for
  Computational Linguistics.

\bibitem[{Clark et~al.(2019)Clark, Khandelwal, Levy, and
  Manning}]{clark-etal-2019-bert}
Kevin Clark, Urvashi Khandelwal, Omer Levy, and Christopher~D. Manning. 2019.
\newblock \href {https://doi.org/10.18653/v1/W19-4828} {What does {BERT} look
  at? an analysis of {BERT}{'}s attention}.
\newblock In \emph{Proceedings of the 2019 ACL Workshop BlackboxNLP: Analyzing
  and Interpreting Neural Networks for NLP}, pages 276--286, Florence, Italy.
  Association for Computational Linguistics.

\bibitem[{Coavoux et~al.(2019)Coavoux, Elsahar, and
  Gall{\'e}}]{coavoux-etal-2019-unsupervised}
Maximin Coavoux, Hady Elsahar, and Matthias Gall{\'e}. 2019.
\newblock \href {https://doi.org/10.18653/v1/D19-5405} {Unsupervised
  aspect-based multi-document abstractive summarization}.
\newblock In \emph{Proceedings of the 2nd Workshop on New Frontiers in
  Summarization}, pages 42--47, Hong Kong, China. Association for Computational
  Linguistics.

\bibitem[{Correia et~al.(2019)Correia, Niculae, and
  Martins}]{correia-etal-2019-adaptively}
Gon{\c{c}}alo~M. Correia, Vlad Niculae, and Andr{\'e} F.~T. Martins. 2019.
\newblock \href {https://doi.org/10.18653/v1/D19-1223} {Adaptively sparse
  transformers}.
\newblock In \emph{Proceedings of the 2019 Conference on Empirical Methods in
  Natural Language Processing and the 9th International Joint Conference on
  Natural Language Processing (EMNLP-IJCNLP)}, pages 2174--2184, Hong Kong,
  China. Association for Computational Linguistics.

\bibitem[{Daume~III and Marcu(2004)}]{daume-iii-marcu-2004-generic}
Hal Daume~III and Daniel Marcu. 2004.
\newblock \href {https://aclanthology.org/W04-1016} {Generic sentence fusion is
  an ill-defined summarization task}.
\newblock In \emph{Text Summarization Branches Out}, pages 96--103, Barcelona,
  Spain. Association for Computational Linguistics.

\bibitem[{Dong et~al.(2019)Dong, Yang, Wang, Wei, Liu, Wang, Gao, Zhou, and
  Hon}]{NIPS2019_9464}
Li~Dong, Nan Yang, Wenhui Wang, Furu Wei, Xiaodong Liu, Yu~Wang, Jianfeng Gao,
  Ming Zhou, and Hsiao-Wuen Hon. 2019.
\newblock Unified language model pre-training for natural language
  understanding and generation.
\newblock In \emph{Advances in Neural Information Processing Systems 32}.

\bibitem[{Erkan and Radev(2004)}]{Erkan:2004}
G\"{u}nes Erkan and Dragomir~R. Radev. 2004.
\newblock \href {https://www.aaai.org/Papers/JAIR/Vol22/JAIR-2214.pdf}
  {{LexRank}: {G}raph-based lexical centrality as salience in text
  summarization}.
\newblock \emph{Journal of Artificial Intelligence Research}.

\bibitem[{Fabbri et~al.(2019)Fabbri, Li, She, Li, and
  Radev}]{fabbri-etal-2019-multi}
Alexander Fabbri, Irene Li, Tianwei She, Suyi Li, and Dragomir Radev. 2019.
\newblock \href {https://doi.org/10.18653/v1/P19-1102} {Multi-news: A
  large-scale multi-document summarization dataset and abstractive hierarchical
  model}.
\newblock In \emph{Proceedings of the 57th Annual Meeting of the Association
  for Computational Linguistics}, pages 1074--1084, Florence, Italy.
  Association for Computational Linguistics.

\bibitem[{Falke and Gurevych(2019)}]{falke-gurevych-2019-fast}
Tobias Falke and Iryna Gurevych. 2019.
\newblock \href {https://doi.org/10.18653/v1/N19-1074} {Fast concept mention
  grouping for concept map-based multi-document summarization}.
\newblock In \emph{Proceedings of the 2019 Conference of the North {A}merican
  Chapter of the Association for Computational Linguistics: Human Language
  Technologies, Volume 1 (Long and Short Papers)}, pages 695--700, Minneapolis,
  Minnesota. Association for Computational Linguistics.

\bibitem[{Gholipour~Ghalandari et~al.(2020)Gholipour~Ghalandari, Hokamp, Pham,
  Glover, and Ifrim}]{gholipour-ghalandari-etal-2020-large}
Demian Gholipour~Ghalandari, Chris Hokamp, Nghia~The Pham, John Glover, and
  Georgiana Ifrim. 2020.
\newblock \href {https://doi.org/10.18653/v1/2020.acl-main.120} {A large-scale
  multi-document summarization dataset from the {W}ikipedia current events
  portal}.
\newblock In \emph{Proceedings of the 58th Annual Meeting of the Association
  for Computational Linguistics}, pages 1302--1308, Online. Association for
  Computational Linguistics.

\bibitem[{Grenander et~al.(2019)Grenander, Dong, Cheung, and
  Louis}]{grenander-etal-2019-countering}
Matt Grenander, Yue Dong, Jackie Chi~Kit Cheung, and Annie Louis. 2019.
\newblock \href {https://doi.org/10.18653/v1/D19-1620} {Countering the effects
  of lead bias in news summarization via multi-stage training and auxiliary
  losses}.
\newblock In \emph{Proceedings of the 2019 Conference on Empirical Methods in
  Natural Language Processing and the 9th International Joint Conference on
  Natural Language Processing (EMNLP-IJCNLP)}, pages 6019--6024, Hong Kong,
  China. Association for Computational Linguistics.

\bibitem[{Haghighi and Vanderwende(2009)}]{haghighi-vanderwende-2009-exploring}
Aria Haghighi and Lucy Vanderwende. 2009.
\newblock \href {https://aclanthology.org/N09-1041} {Exploring content models
  for multi-document summarization}.
\newblock In \emph{Proceedings of Human Language Technologies: The 2009 Annual
  Conference of the North {A}merican Chapter of the Association for
  Computational Linguistics}, pages 362--370, Boulder, Colorado. Association
  for Computational Linguistics.

\bibitem[{Hamilton(2014)}]{Hamilton:2014}
Jillian Hamilton.
\newblock \href
  {https://news.clearancejobs.com/2014/10/30/day-life-intelligence-analyst/} {A
  day in the life of an intelligence analyst} [online]. 2014.

\bibitem[{Handler and O{'}Connor(2018)}]{handler-oconnor-2018-relational}
Abram Handler and Brendan O{'}Connor. 2018.
\newblock \href {https://doi.org/10.18653/v1/N18-1159} {Relational
  summarization for corpus analysis}.
\newblock In \emph{Proceedings of the 2018 Conference of the North {A}merican
  Chapter of the Association for Computational Linguistics: Human Language
  Technologies, Volume 1 (Long Papers)}, pages 1760--1769, New Orleans,
  Louisiana. Association for Computational Linguistics.

\bibitem[{Hintzman(1976)}]{Hintzman:1976}
Douglas~L. Hintzman. 1976.
\newblock Repetition and memory.
\newblock \emph{Psychology of Learning and Motivation}, 10:47--91.

\bibitem[{Hong et~al.(2014)Hong, Conroy, Favre, Kulesza, Lin, and
  Nenkova}]{Hong:2014}
Kai Hong, John~M Conroy, Benoit Favre, Alex Kulesza, Hui Lin, and Ani Nenkova.
  2014.
\newblock \href
  {http://www.lrec-conf.org/proceedings/lrec2014/pdf/1093_Paper.pdf} {A
  repository of state of the art and competitive baseline summaries for generic
  news summarization}.
\newblock In \emph{Proceedings of the Ninth International Conference on
  Language Resources and Evaluation (LREC)}.

\bibitem[{Huang et~al.(2021)Huang, Cao, Parulian, Ji, and
  Wang}]{huang-etal-2021-efficient}
Luyang Huang, Shuyang Cao, Nikolaus Parulian, Heng Ji, and Lu~Wang. 2021.
\newblock \href {https://doi.org/10.18653/v1/2021.naacl-main.112} {Efficient
  attentions for long document summarization}.
\newblock In \emph{Proceedings of the 2021 Conference of the North American
  Chapter of the Association for Computational Linguistics: Human Language
  Technologies}, pages 1419--1436, Online. Association for Computational
  Linguistics.

\bibitem[{Jin and Wan(2020)}]{jin-wan-2020-abstractive}
Hanqi Jin and Xiaojun Wan. 2020.
\newblock \href {https://doi.org/10.18653/v1/2020.findings-emnlp.231}
  {Abstractive multi-document summarization via joint learning with
  single-document summarization}.
\newblock In \emph{Findings of the Association for Computational Linguistics:
  EMNLP 2020}, pages 2545--2554, Online. Association for Computational
  Linguistics.

\bibitem[{Kingma and Ba(2015)}]{Kingma:2015}
Diederik~P. Kingma and Jimmy Ba. 2015.
\newblock Adam: {A} method for stochastic optimization.
\newblock In \emph{Proceedings of the International Conference on Learning
  Representations (ICLR)}.

\bibitem[{Kryscinski et~al.(2019)Kryscinski, Keskar, McCann, Xiong, and
  Socher}]{kryscinski-etal-2019-neural}
Wojciech Kryscinski, Nitish~Shirish Keskar, Bryan McCann, Caiming Xiong, and
  Richard Socher. 2019.
\newblock \href {https://doi.org/10.18653/v1/D19-1051} {Neural text
  summarization: A critical evaluation}.
\newblock In \emph{Proceedings of the 2019 Conference on Empirical Methods in
  Natural Language Processing and the 9th International Joint Conference on
  Natural Language Processing (EMNLP-IJCNLP)}, pages 540--551, Hong Kong,
  China. Association for Computational Linguistics.

\bibitem[{Kulesza and Taskar(2012)}]{Kulesza:2012}
Alex Kulesza and Ben Taskar. 2012.
\newblock \href {https://arxiv.org/abs/1207.6083} {\emph{Determinantal Point
  Processes for Machine Learning}}.
\newblock Now Publishers Inc.

\bibitem[{Laban et~al.(2020)Laban, Hsi, Canny, and
  Hearst}]{laban-etal-2020-summary}
Philippe Laban, Andrew Hsi, John Canny, and Marti~A. Hearst. 2020.
\newblock \href {https://doi.org/10.18653/v1/2020.acl-main.460} {The summary
  loop: Learning to write abstractive summaries without examples}.
\newblock In \emph{Proceedings of the 58th Annual Meeting of the Association
  for Computational Linguistics}, pages 5135--5150, Online. Association for
  Computational Linguistics.

\bibitem[{Lebanoff et~al.(2018)Lebanoff, Song, and
  Liu}]{lebanoff-etal-2018-adapting}
Logan Lebanoff, Kaiqiang Song, and Fei Liu. 2018.
\newblock \href {https://doi.org/10.18653/v1/D18-1446} {Adapting the neural
  encoder-decoder framework from single to multi-document summarization}.
\newblock In \emph{Proceedings of the 2018 Conference on Empirical Methods in
  Natural Language Processing}, pages 4131--4141, Brussels, Belgium.
  Association for Computational Linguistics.

\bibitem[{Lewis et~al.(2020)Lewis, Liu, Goyal, Ghazvininejad, Mohamed, Levy,
  Stoyanov, and Zettlemoyer}]{lewis-etal-2020-bart}
Mike Lewis, Yinhan Liu, Naman Goyal, Marjan Ghazvininejad, Abdelrahman Mohamed,
  Omer Levy, Veselin Stoyanov, and Luke Zettlemoyer. 2020.
\newblock \href {https://doi.org/10.18653/v1/2020.acl-main.703} {{BART}:
  Denoising sequence-to-sequence pre-training for natural language generation,
  translation, and comprehension}.
\newblock In \emph{Proceedings of the 58th Annual Meeting of the Association
  for Computational Linguistics}, pages 7871--7880, Online. Association for
  Computational Linguistics.

\bibitem[{Li et~al.(2020)Li, Maskharashvili, Jory Stevens-Guille, and
  White}]{li-etal-2020-leveraging}
Xintong Li, Aleksandre Maskharashvili, Symon Jory Stevens-Guille, and Michael
  White. 2020.
\newblock \href {https://aclanthology.org/2020.webnlg-1.12} {Leveraging large
  pretrained models for {W}eb{NLG} 2020}.
\newblock In \emph{Proceedings of the 3rd International Workshop on Natural
  Language Generation from the Semantic Web (WebNLG+)}, pages 117--124, Dublin,
  Ireland (Virtual). Association for Computational Linguistics.

\bibitem[{Lin(2004)}]{lin-2004-rouge}
Chin-Yew Lin. 2004.
\newblock \href {https://aclanthology.org/W04-1013} {{ROUGE}: A package for
  automatic evaluation of summaries}.
\newblock In \emph{Text Summarization Branches Out}, pages 74--81, Barcelona,
  Spain. Association for Computational Linguistics.

\bibitem[{Lin and Bilmes(2011)}]{lin-bilmes-2011-class}
Hui Lin and Jeff Bilmes. 2011.
\newblock \href {https://aclanthology.org/P11-1052} {A class of submodular
  functions for document summarization}.
\newblock In \emph{Proceedings of the 49th Annual Meeting of the Association
  for Computational Linguistics: Human Language Technologies}, pages 510--520,
  Portland, Oregon, USA. Association for Computational Linguistics.

\bibitem[{Liu and Lapata(2019)}]{liu-lapata-2019-hierarchical}
Yang Liu and Mirella Lapata. 2019.
\newblock \href {https://doi.org/10.18653/v1/P19-1500} {Hierarchical
  transformers for multi-document summarization}.
\newblock In \emph{Proceedings of the 57th Annual Meeting of the Association
  for Computational Linguistics}, pages 5070--5081, Florence, Italy.
  Association for Computational Linguistics.

\bibitem[{Mao et~al.(2020)Mao, Qu, Xie, Ren, and Han}]{mao-etal-2020-multi}
Yuning Mao, Yanru Qu, Yiqing Xie, Xiang Ren, and Jiawei Han. 2020.
\newblock \href {https://doi.org/10.18653/v1/2020.emnlp-main.136}
  {Multi-document summarization with maximal marginal relevance-guided
  reinforcement learning}.
\newblock In \emph{Proceedings of the 2020 Conference on Empirical Methods in
  Natural Language Processing (EMNLP)}, pages 1737--1751, Online. Association
  for Computational Linguistics.

\bibitem[{Mihalcea and Tarau(2004)}]{mihalcea-tarau-2004-textrank}
Rada Mihalcea and Paul Tarau. 2004.
\newblock \href {https://aclanthology.org/W04-3252} {{T}ext{R}ank: Bringing
  order into text}.
\newblock In \emph{Proceedings of the 2004 Conference on Empirical Methods in
  Natural Language Processing}, pages 404--411, Barcelona, Spain. Association
  for Computational Linguistics.

\bibitem[{Miller(2020)}]{Miller:2020}
Ron Miller. 2020.
\newblock Former salesforce chief scientist announces new search engine to take
  on google.
\newblock
  \emph{https://techcrunch.com/2020/12/08/former-salesforce-chief-scientist-announces-new-search-engine-to-take-on-google/}.

\bibitem[{Narayan et~al.(2018)Narayan, Cohen, and
  Lapata}]{narayan-etal-2018-dont}
Shashi Narayan, Shay~B. Cohen, and Mirella Lapata. 2018.
\newblock \href {https://doi.org/10.18653/v1/D18-1206} {Don{'}t give me the
  details, just the summary! topic-aware convolutional neural networks for
  extreme summarization}.
\newblock In \emph{Proceedings of the 2018 Conference on Empirical Methods in
  Natural Language Processing}, pages 1797--1807, Brussels, Belgium.
  Association for Computational Linguistics.

\bibitem[{Nayeem et~al.(2018)Nayeem, Fuad, and
  Chali}]{nayeem-etal-2018-abstractive}
Mir~Tafseer Nayeem, Tanvir~Ahmed Fuad, and Yllias Chali. 2018.
\newblock \href {https://aclanthology.org/C18-1102} {Abstractive unsupervised
  multi-document summarization using paraphrastic sentence fusion}.
\newblock In \emph{Proceedings of the 27th International Conference on
  Computational Linguistics}, pages 1191--1204, Santa Fe, New Mexico, USA.
  Association for Computational Linguistics.

\bibitem[{Nenkova and Passonneau(2004)}]{nenkova-passonneau-2004-evaluating}
Ani Nenkova and Rebecca Passonneau. 2004.
\newblock \href {https://aclanthology.org/N04-1019} {Evaluating content
  selection in summarization: The pyramid method}.
\newblock In \emph{Proceedings of the Human Language Technology Conference of
  the North {A}merican Chapter of the Association for Computational
  Linguistics: {HLT}-{NAACL} 2004}, pages 145--152, Boston, Massachusetts, USA.
  Association for Computational Linguistics.

\bibitem[{Perez-Beltrachini et~al.(2019)Perez-Beltrachini, Liu, and
  Lapata}]{perez-beltrachini-etal-2019-generating}
Laura Perez-Beltrachini, Yang Liu, and Mirella Lapata. 2019.
\newblock \href {https://doi.org/10.18653/v1/P19-1504} {Generating summaries
  with topic templates and structured convolutional decoders}.
\newblock In \emph{Proceedings of the 57th Annual Meeting of the Association
  for Computational Linguistics}, pages 5107--5116, Florence, Italy.
  Association for Computational Linguistics.

\bibitem[{Radford et~al.(2019)Radford, Wu, Child, Luan, Amodei, and
  Sutskever}]{radford2019language}
Alec Radford, Jeff Wu, Rewon Child, David Luan, Dario Amodei, and Ilya
  Sutskever. 2019.
\newblock \href
  {https://d4mucfpksywv.cloudfront.net/better-language-models/language_models_are_unsupervised_multitask_learners.pdf}
  {Language models are unsupervised multitask learners}.

\bibitem[{Raffel et~al.(2020)Raffel, Shazeer, Roberts, Lee, Narang, Matena,
  Zhou, Li, and Liu}]{JMLR:v21:20-074}
Colin Raffel, Noam Shazeer, Adam Roberts, Katherine Lee, Sharan Narang, Michael
  Matena, Yanqi Zhou, Wei Li, and Peter~J. Liu. 2020.
\newblock \href {http://jmlr.org/papers/v21/20-074.html} {Exploring the limits
  of transfer learning with a unified text-to-text transformer}.
\newblock \emph{Journal of Machine Learning Research}, 21(140):1--67.

\bibitem[{Rogers et~al.(2020)Rogers, Kovaleva, and
  Rumshisky}]{rogers2020primer}
Anna Rogers, Olga Kovaleva, and Anna Rumshisky. 2020.
\newblock \href {http://arxiv.org/abs/2002.12327} {A primer in bertology: What
  we know about how bert works}.

\bibitem[{See et~al.(2017)See, Liu, and Manning}]{see-etal-2017-get}
Abigail See, Peter~J. Liu, and Christopher~D. Manning. 2017.
\newblock \href {https://doi.org/10.18653/v1/P17-1099} {Get to the point:
  Summarization with pointer-generator networks}.
\newblock In \emph{Proceedings of the 55th Annual Meeting of the Association
  for Computational Linguistics (Volume 1: Long Papers)}, pages 1073--1083,
  Vancouver, Canada. Association for Computational Linguistics.

\bibitem[{Vaswani et~al.(2017)Vaswani, Shazeer, Parmar, Uszkoreit, Jones,
  Gomez, Kaiser, and Polosukhin}]{NIPS2017_7181}
Ashish Vaswani, Noam Shazeer, Niki Parmar, Jakob Uszkoreit, Llion Jones,
  Aidan~N Gomez, \L~ukasz Kaiser, and Illia Polosukhin. 2017.
\newblock Attention is all you need.
\newblock In I.~Guyon, U.~V. Luxburg, S.~Bengio, H.~Wallach, R.~Fergus,
  S.~Vishwanathan, and R.~Garnett, editors, \emph{Advances in Neural
  Information Processing Systems 30}.

\bibitem[{Xing et~al.(2021)Xing, Xiao, and Carenini}]{xing-etal-2021-demoting}
Linzi Xing, Wen Xiao, and Giuseppe Carenini. 2021.
\newblock \href {https://doi.org/10.18653/v1/2021.acl-short.119} {Demoting the
  lead bias in news summarization via alternating adversarial learning}.
\newblock In \emph{Proceedings of the 59th Annual Meeting of the Association
  for Computational Linguistics and the 11th International Joint Conference on
  Natural Language Processing (Volume 2: Short Papers)}, pages 948--954,
  Online. Association for Computational Linguistics.

\bibitem[{Xu and Durrett(2021)}]{xu-durrett-2021-dissecting}
Jiacheng Xu and Greg Durrett. 2021.
\newblock \href {https://doi.org/10.18653/v1/2021.acl-long.539} {Dissecting
  generation modes for abstractive summarization models via ablation and
  attribution}.
\newblock In \emph{Proceedings of the 59th Annual Meeting of the Association
  for Computational Linguistics and the 11th International Joint Conference on
  Natural Language Processing (Volume 1: Long Papers)}, pages 6925--6940,
  Online. Association for Computational Linguistics.

\bibitem[{Zhang et~al.(2020{\natexlab{a}})Zhang, Zhao, Saleh, and
  Liu}]{pmlr-v119-zhang20ae}
Jingqing Zhang, Yao Zhao, Mohammad Saleh, and Peter Liu. 2020{\natexlab{a}}.
\newblock \href {https://proceedings.mlr.press/v119/zhang20ae.html} {{PEGASUS}:
  Pre-training with extracted gap-sentences for abstractive summarization}.
\newblock In \emph{Proceedings of the 37th International Conference on Machine
  Learning}, volume 119 of \emph{Proceedings of Machine Learning Research},
  pages 11328--11339. PMLR.

\bibitem[{Zhang et~al.(2020{\natexlab{b}})Zhang, Kishore, Wu, Weinberger, and
  Artzi}]{Zhang2020BERTScore}
Tianyi Zhang, Varsha Kishore, Felix Wu, Kilian~Q. Weinberger, and Yoav Artzi.
  2020{\natexlab{b}}.
\newblock \href {https://openreview.net/forum?id=SkeHuCVFDr} {Bertscore:
  Evaluating text generation with bert}.
\newblock In \emph{International Conference on Learning Representations}.

\bibitem[{Zhang et~al.(2019)Zhang, Wei, and Zhou}]{zhang-etal-2019-hibert}
Xingxing Zhang, Furu Wei, and Ming Zhou. 2019.
\newblock \href {https://doi.org/10.18653/v1/P19-1499} {{HIBERT}: Document
  level pre-training of hierarchical bidirectional transformers for document
  summarization}.
\newblock In \emph{Proceedings of the 57th Annual Meeting of the Association
  for Computational Linguistics}, pages 5059--5069, Florence, Italy.
  Association for Computational Linguistics.

\bibitem[{Zhao et~al.(2019)Zhao, Peyrard, Liu, Gao, Meyer, and
  Eger}]{zhao-etal-2019-moverscore}
Wei Zhao, Maxime Peyrard, Fei Liu, Yang Gao, Christian~M. Meyer, and Steffen
  Eger. 2019.
\newblock \href {https://doi.org/10.18653/v1/D19-1053} {{M}over{S}core: Text
  generation evaluating with contextualized embeddings and earth mover
  distance}.
\newblock In \emph{Proceedings of the 2019 Conference on Empirical Methods in
  Natural Language Processing and the 9th International Joint Conference on
  Natural Language Processing (EMNLP-IJCNLP)}, pages 563--578, Hong Kong,
  China. Association for Computational Linguistics.

\end{thebibliography}

\end{document}